\pdfoutput=1

\documentclass[11pt]{article}

\usepackage[]{EMNLP2022}

\usepackage{times}
\usepackage{latexsym}
\usepackage{graphicx}
\usepackage{enumitem}
\usepackage{algorithm}
\usepackage{algpseudocode}
\usepackage{amssymb}
\usepackage{multirow}
\usepackage{booktabs}
\usepackage{mathtools}
\usepackage{amsfonts}
\usepackage{amsmath}
\usepackage{hyperref}
\usepackage{bm}

\usepackage[T1]{fontenc}

\usepackage[utf8]{inputenc}

\usepackage{microtype}

\usepackage{inconsolata}

\title{Text Editing as Imitation Game}

\author{\thanks{$^{*}$ Work was done at Alibaba Group.}$\;\,$Ning Shi$^{\spadesuit\heartsuit}$\qquad Bin Tang$^{\heartsuit}$\qquad Bo Yuan$^{\heartsuit}$\qquad Longtao Huang$^{\heartsuit}$\qquad \\ \textbf{Yewen Pu$^{\clubsuit}$\qquad Jie Fu$^{\diamondsuit}$\qquad \thanks{$^{*}$ Zhouhan Lin is the corresponding author.}$\;\,$Zhouhan Lin$^{\bigstar}$} \\
  $^{\spadesuit}$Alberta Machine Intelligence Institute, Dept. of Computing Science, University of Alberta \\
  $^{\heartsuit}$Alibaba Group\qquad $^{\bigstar}$Shanghai Jiao Tong University \\
  $^{\clubsuit}$Autodesk Research\qquad $^{\diamondsuit}$Beijing Academy of Artificial Intelligence \\ 
  \texttt{ning.shi@ualberta.ca, \{tangbin.tang,qiufu.yb,kaiyang.hlt\}@alibaba-inc.com} \\
  \texttt{yewen.pu@autodesk.com, fujie@baai.ac.cn, lin.zhouhan@gmail.com} \\}

\begin{document}
\maketitle

\begin{abstract}
Text editing, such as grammatical error correction, arises naturally from imperfect textual data. Recent works frame text editing as a multi-round sequence tagging task, where operations -- such as insertion and substitution -- are represented as a sequence of tags. While achieving good results, this encoding is limited in flexibility as all actions are bound to token-level tags. In this work, we reformulate text editing as an imitation game using behavioral cloning. Specifically, we convert conventional sequence-to-sequence data into state-to-action demonstrations, where the action space can be as flexible as needed. Instead of generating the actions one at a time, we introduce a dual decoders structure to parallel the decoding while retaining the dependencies between action tokens, coupled with trajectory augmentation to alleviate the distribution shift that imitation learning often suffers. In experiments on a suite of Arithmetic Equation benchmarks, our model consistently outperforms the autoregressive baselines in terms of performance, efficiency, and robustness. We hope our findings will shed light on future studies in reinforcement learning applying sequence-level action generation to natural language processing.
\end{abstract}

\section{Introduction}

\begin{figure}[t!]
    \centering
    \includegraphics[width=\linewidth]{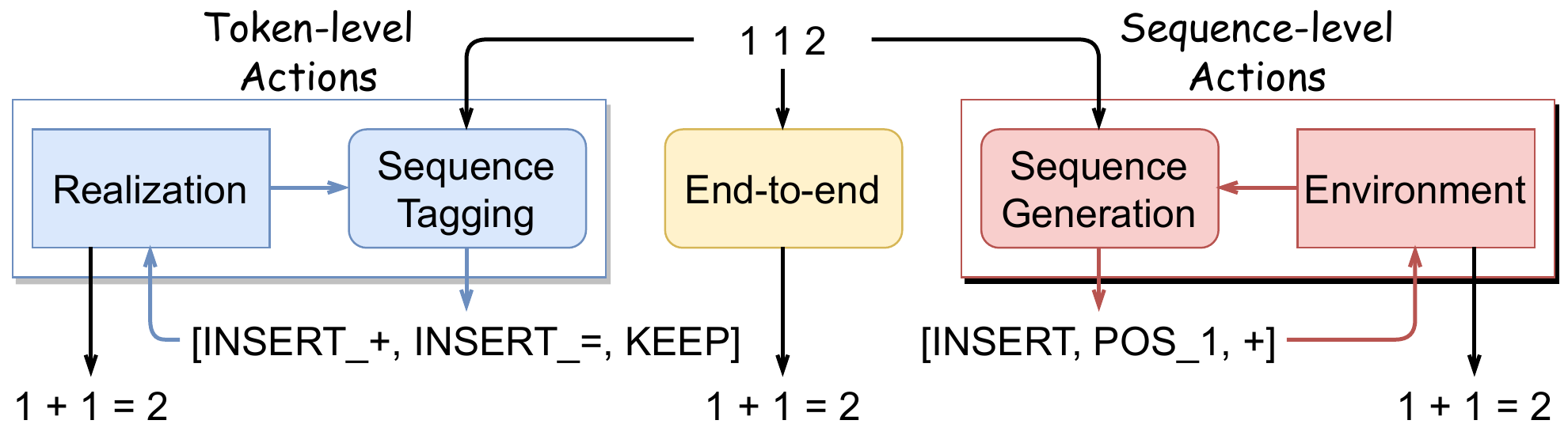}
    \vspace{-6mm}
    \caption{Three approaches -- sequence tagging (left), end-to-end (middle), sequence generation (right) -- to turn an invalid arithmetic expression ``1 1 2'' into a valid one ``1 + 1 = 2''. In end-to-end, the entire string ``1 1 2'' is encoded into a latent state, which the string ``1 + 1 = 2'' is generated directly. In sequence tagging, a localized action (such as ``INSERT\_+'', meaning insert a ``+'' symbol after this token) is applied/tagged to each token; these token-level actions are then executed, modifying the input string. In contrast, sequence generation output an entire action sequence, generating the location (rather than tagging it), and the action sequence is executed, modifying the input string. Both token-level actions and sequence-level actions can be applied multiple times to polish the text further (up to a fixed point).}
    \label{figure:general}
    \vspace{-2mm}
\end{figure}

Text editing \citep{malmi-etal-2022-text} is an important domain of processing tasks to edit the text in a localized fashion, applying to text simplification \citep{agrawal-etal-2021-non}, grammatical error correction \citep{li2022sequence}, punctuation restoration \citep{shi21_interspeech}, to name a few. Neural sequence-to-sequence (seq2seq) framework \citep{sutskever2014sequence} establishes itself as the primary approach to text editing tasks, by framing the problem as machine translation \cite{wu2016google}. Applying a seq2seq modeling has the advantage of simplicity, where the system can simply be built by giving input-output pairs consisting of pathological sequences to be edited, and the desired sequence output, without much manual processing efforts \cite{junczys-dowmunt-etal-2018-approaching}.

However, even with a copy mechanism \citep{see-etal-2017-get,zhao-etal-2019-improving,Panthaplackel_Allamanis_Brockschmidt_2021}, an end-to-end model can struggle in carrying out localized, specific fixes while keeping the rest of the sequence intact. Thus, sequence tagging is often found more appropriate when outputs highly overlap with inputs \cite{dong-etal-2019-editnts,mallinson-etal-2020-felix,stahlberg-kumar-2020-seq2edits}. In such cases, a neural model predicts a tag sequence -- representing localized fixes such as insertion and substitution -- and a programmatic interpreter implements these edit operations through. Here, each tag represents a \emph{token-level} action and determines the operation on its attached token \citep{kohita-etal-2020-q}. A model can avoid modifying the overlap by assigning no-op (e.g., \texttt{KEEP}), while the action space is limited to token-level modifications, such as deletion or insertion after a token \citep{awasthi-etal-2019-parallel,malmi-etal-2019-encode}.

In contrast, alternative approaches \citep{Gupta_Kanade_Shevade_2019} train the agent to explicitly generate free-form edit actions and iteratively reconstructs the text during the interaction with an environment capable of altering the text based on these actions. This \emph{sequence-level} action generation \citep{branavan-etal-2009-reinforcement,guu-etal-2017-language,elgohary-etal-2021-nl} allows higher flexibility of action design not limited to token-level actions, and is more advantageous given the narrowed problem space and dynamic context in the edit \citep{shi-etal-2020-recurrent}.

The mechanisms of sequence tagging and sequence generation against end-to-end are exemplified in Figure \ref{figure:general}. Both methods allow multiple rounds of sequence refinement \citep{ge-etal-2018-fluency,liu-etal-2021-learning-ask} and imitation learning (IL) \citep{10.1162/neco.1991.3.1.88}. Essentially an agent learns from the demonstrations of an expert policy and later imitates the memorized behavior to act independently \citep{NIPS1996_68d13cf2}. On the one hand, IL in sequence tagging functions as a standard supervised learning in its nature and thus has attracted significant interest and been widely used recently \citep{agrawal-etal-2021-non,yao2021learning,agrawal-carpuat-2022-imitation}, achieving good results in the token-level action generation setting \citep{levenshtein-transformer,reid-zhong-2021-lewis}. On the other hand, IL in sequence-level action generation is less well defined even though its principle has been followed in text editing \citep{shi-etal-2020-recurrent} and many others \citep{chen-etal-2021-multi}. As a major obstacle, the training is on state-action demonstrations, where the encoding of the states and actions can be very different \citep{gu2018nonautoregressive}. For instance, the mismatch of the lengths dimension between the state and action makes it tricky to implement for an auto-regressive modeling that benefits from a single, uniform representation.

To tackle the issues above, we reformulate text editing as an \emph{imitation game} controlled by a Markov Decision Process (MDP). To begin with, we define the input sequence as the initial state, the required operations as action sequences, and the output target sequence as the goal state. A learning agent needs to imitate an expert policy, respond to seen states with actions, and interact with the environment until the success of the eventual editing. To convert existing input-output data into state-action pairs, we utilize \emph{trajectory generation} (TG), a skill to leverage dynamic programming (DP) for an efficient search of the minimum operations given a predefined edit metric. We backtrace explored editing paths and automatically express operations as action sequences. Regarding the length misalignment, we first take advantage of the flexibility at the sequence-level to fix actions to be of the same length. Secondly, we employ a linear layer after the encoder to transform the length dimension of the context matrix into the action length. By that, we introduce a \emph{dual decoders} (D2) structure that not only parallels the decoding but also retains capturing interdependencies among action tokens. Taking a further step, we propose \emph{trajectory augmentation} (TA) as a solution to the distribution shift problem most IL suffers \citep{ross2011reduction}. Through a suite of three Arithmetic Equation (AE) benchmarks \citep{shi-etal-2020-recurrent}, namely Arithmetic Operators Restoration (AOR), Arithmetic Equation Simplification (AES), and Arithmetic Equation Correction (AEC), we confirm the superiority of our learning paradigm. In particular, D2 consistently exceeds standard autoregressive models from performance, efficiency, and robustness perspectives.

In theory, our methods also apply to other imitation learning scenarios where a reward function exists to further promote the agent. In this work, we primarily focus on a proof-of-concept of our learning paradigm landing at supervised behavior cloning (BC) in the context of text editing. To this end, our contributions\footnote{Code and data are publicly available at \href{https://github.com/ShiningLab/Text-Editing-as-Imitation-Game}{GitHub}.} are as follows:
\begin{enumerate}[leftmargin=*,noitemsep,topsep=0pt]
    \item We frame text editing into an imitation game formally defined as an MDP, allowing the highest degrees of flexibility to design actions at the sequence-level.
    \item We involve TG to translate input-output data to state-action demonstrations for IL.
    \item We introduce D2, a novel non-autoregressive decoder, boosting the learning in terms of accuracy, efficiency, and robustness.
    \item We propose a corresponding TA technique to mitigate distribution shift IL often suffers.
\end{enumerate}

\begin{figure*}[t]
    \centering
    \includegraphics[width=\textwidth]{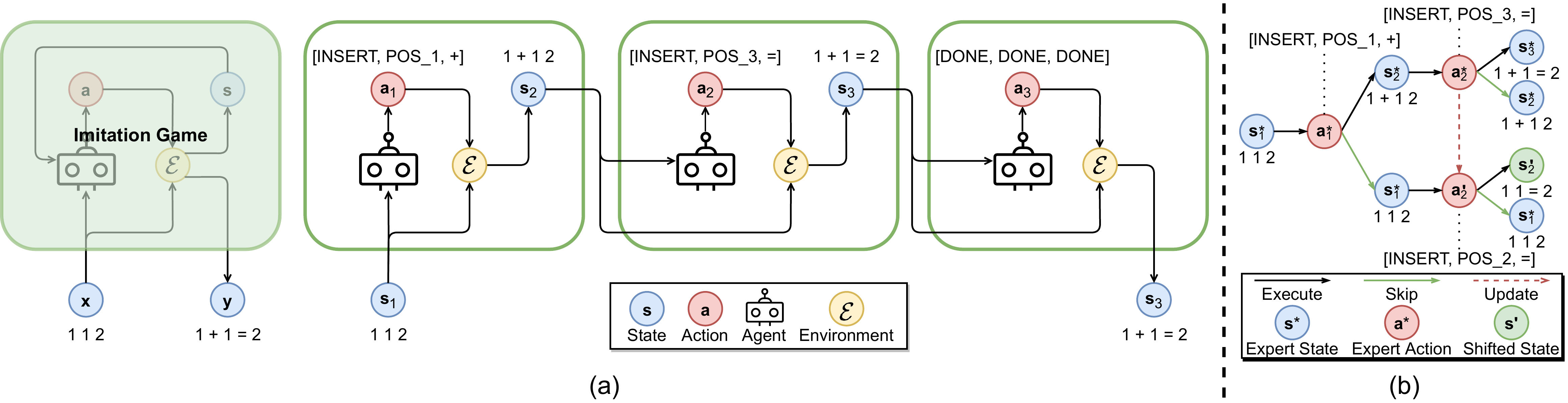}
    \vspace{-6mm}
    \caption{(a) shows the imitation game of AOR. Considering input text $\mathbf{x}$ as initial state $\mathbf{s}_{1}$, the agent interacts with the environment to edit ``$1\;1\;2$'' into ``$1 + 1 = 2$'' via action $\mathbf{a}_{1}$ to insert ``$+$'' at the first position and $\mathbf{a}_{2}$ to insert ``$=$'' at the thrid position. After $\mathbf{a}_{3}$, the agent stops editing and calls the environment to return $\mathbf{s}_{3}$ as the output text $\mathbf{y}$. Using the same example, (b) explains how to achieve shifted state $\mathbf{s}^{\prime}_{2}$ by skipping action $\mathbf{a}^{*}_{1}$ and doing $\mathbf{a}^{\prime}_{2}$. Here we update $\mathbf{a}^{*}_{2}$ to $\mathbf{a}^{\prime}_{2}$ accordingly due to the previous skipping. The new state $\mathbf{s}^{\prime}_{2}$ was not in the expert demonstrations.}
    \label{figure:method}
    \vspace{-2mm}
\end{figure*}

\section{Imitation Game}

We aim to cast text editing into an imitation game by defining the task as a recurrent sequence generation, as presented in Figure \ref{figure:method} (a). In this section, we describe the major components of our proposal, including (1) the problem definition, (2) the data translation, (3) the model structure, and (4) a solution to the distribution shift.

\subsection{Behavior cloning}

\begin{figure*}[t]
    \centering
    \includegraphics[width=\linewidth]{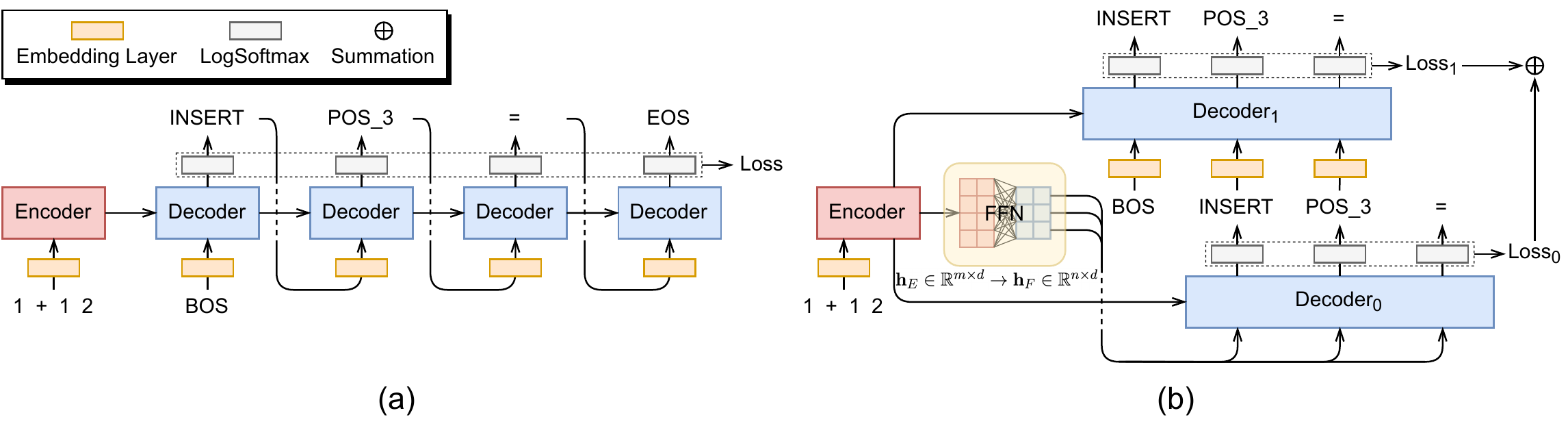}
    \vspace{-6mm}
    \caption{The conventional autoregressive decoder (a) compared with the proposed non-autoregressive D2 (b) in which the linear layer aligns the sequence length dimension for the subsequent parallel decoding.}
    \label{figure:model}
    \vspace{-2mm}
\end{figure*}

We tear a text editing task $\mathcal{X} \mapsto \mathcal{Y}$ into recurrent subtasks of sequence generation $\mathcal{S} \mapsto \mathcal{A}$ defined by an MDP tuple $\mathcal{M}=(\mathcal{S}, \mathcal{A}, \mathcal{P}, \mathcal{E}, \mathcal{R})$.
\\
\textbf{State} $\mathcal{S}$ is a set of text sequences $\mathbf{s} = s_{j\leq m}$, where $s \in \mathcal{V}_{\mathcal{S}}$. We think of a source sequence $\mathbf{x} \in \mathcal{X}$ as the initial state $\mathbf{s}_{1}$, its target sequence $\mathbf{y} \in \mathcal{Y}$ as the goal state $\mathbf{s}_{T}$, and every edited sequence in between as an intermediate state $\mathbf{s}_{t}$. The path $\mathbf{x} \mapsto \mathbf{y}$ can be represented as a set of sequential states $\mathbf{s}_{t\leq T}$.
\\
\textbf{Action} $\mathcal{A}$ is a set of action sequences $\mathbf{a} = a_{i\leq n}$, where $a \in \mathcal{V}_{\mathcal{A}}$. In Figure \ref{figure:model}, ``INSERT'', ``POS\_3'', and ``='' are three action tokens belonging to the vocabulary space of action $\mathcal{V}_{\mathcal{A}}$. In contrast to token-level actions in sequence tagging, sentence-level ones set free the editing by varying edit metrics $\mathbf{E}$ (e.g., Levenshtein distance) as long as $\mathcal{X}\xmapsto[]{\mathcal{A}_{\mathbf{E}}}\mathcal{Y}$. It serves as an expert policy $\pi^{*}$ to demonstrate the path to the goal state. A better expert usually means better demonstrations and imitation results. Hence, depending on the task, a suitable $\mathbf{E}$ is essential.
\\
\textbf{Transition matrix} $\mathcal{P}$ models the probability $p$ that an action $\mathbf{a}_{t}$ leads a state $\mathbf{s}_{t}$ to the state $\mathbf{s}_{t+1}$. We know $\forall \mathbf{s}, \mathbf{a}.\;p(\mathbf{s}_{t+1}|\mathbf{s}_{t},\mathbf{a}_{t})=1$ due to the nature of text editing. So we can omit $\mathcal{P}$.
\\
\textbf{Environment} $\mathcal{E}$ responds to an action and updates the game state accordingly by $\mathbf{s}_{t+1}=\mathcal{E}(\mathbf{s}_{t}, \mathbf{a}_{t})$ with process control. For example, the environment can refuse to execute actions that fail to pass the verification and terminate the game if a maximum number of iterations has been consumed. 
\\
\textbf{Reward function} $\mathcal{R}$ calculates a reward for each action. It is a major factor contributing to the success of reinforcement learning. In the scope of this paper, we focus on BC, the simplest form of IL. So we can also omit $\mathcal{R}$ and leave it for future work. 

The formulation turns out to be a simplified $\mathcal{M}_{BC}=(\mathcal{S}, \mathcal{A}, \mathcal{E})$. Interacting with the environment $\mathcal{E}$, we hope a trained agent is able to follow its learned policy $\pi:\mathcal{S} \mapsto \mathcal{A}$, and iteratively edit the initial state $\mathbf{s}_{0}=\mathbf{x}$ into the goal state $\mathbf{s}_{T}=\mathbf{y}$.

\subsection{Trajectory generation}

\begin{algorithm}[t]
    \small
    \caption{Trajectory Generation ($\mathrm{TG}$)}
    \label{alg:traj_gen}
    \renewcommand{\algorithmicrequire}{\textbf{Input:}}
    \renewcommand{\algorithmicensure}{\textbf{Output:}}
    \begin{algorithmic}[1]
        \Require
            Initial state $\mathbf{x}$, goal state $\mathbf{y}$, environment $\mathcal{E}$, and edit metric $\mathbf{E}$.
        \Ensure
            Trajectories $\tau$.
        \State $\tau \gets \emptyset$
        \State $\mathbf{s} \gets \mathbf{x}$
        \State $ops \gets \mathrm{DP}(\mathbf{x},\mathbf{y}, E)$
        \For{$op \in ops$}
            \State $\mathbf{a} \gets$ $\mathrm{Action}(op)$ \Comment{Translate operation to action}
            \State $\tau \gets \tau \cup [(\mathbf{s}, \mathbf{a})]$
            \State $\mathbf{s} \gets \mathcal{E}(\mathbf{s}, \mathbf{a}$)
        \EndFor
        \State $\tau \gets \tau \cup [(\mathbf{s}, \mathbf{a}_{T})]$ \Comment{Append goal state and output action}
        \State \Return $\tau$
    \end{algorithmic}
\end{algorithm}

A data set to learn $\mathcal{X} \mapsto \mathcal{Y}$ consists of input-output pairs. It is necessary to convert it into state-action ones so that an agent can mimic the expert policy $\pi^{*}:\mathcal{S} \mapsto \mathcal{A}$ via supervised learning. A detailed TG is described in Algorithm \ref{alg:traj_gen}.

Treating a pre-defined edit metric $\mathbf{E}$ as the expert policy $\pi^{*}$, we can leverage DP to efficiently find the minimum operations required to convert $\mathbf{x}$ into $\mathbf{y}$ in a left-to-right manner and backtrace this path to get specific operations.

Operations are later expressed as a set of sequential actions $\mathbf{a}^{*}_{t\leq T}$. Here we utilize a special symbol \texttt{DONE} to mark the last action $\mathbf{a}^{*}_{T}$ where $\forall a \in \mathbf{a}^{*}_{T}.\;a=\texttt{DONE}$. Once an agent performs $\mathbf{a}^{*}_{T}$, the current state is returned by the environment as the final output.

Given $\mathbf{s}^{*}_{1}=\mathbf{x}$, we attain the next state $\mathbf{s}^{*}_{2}=\mathcal{E}(\mathbf{s}^{*}_{1}, \mathbf{a}^{*}_{1})$ and continue the rest until achieving $\mathbf{s}^{*}_{T}=\mathbf{y}$, resulting in a set of sequential states $\mathbf{s}^{*}_{t\leq T}$.

After one-to-one correspondence between states and actions, we collect a set of sequential expert's demonstrations $\tau^{*}=[(\mathbf{s}^{*}_{t\leq T}, \mathbf{a}^{*}_{t\leq T})]$. Repeating the same process, we eventually convert $\mathcal{X} \mapsto \mathcal{Y}$ into trajectories $\mathcal{T}^{*}: \mathcal{S} \mapsto \mathcal{A}$.

\subsection{Model architecture}

We form $\mathcal{S} \mapsto \mathcal{A}$ as sequence generation. More precisely, a neural model (i.e., the agent) takes states as input and outputs actions. Training an imitation policy with BC corresponds to fitting a parametric model $\pi_{\theta}$ that minimizes the negative log-likelihood loss $l(\mathbf{a}^{*}, \pi_{\theta}(\mathbf{s}))$. Most seq2seq models have an encoder-decoder structure. 
\\
\textbf{Encoder} takes an embedded state $\mathrm{E}(\mathbf{s})\in\mathbb{R}^{m\times d}$ and generates an encoded hidden state $\mathbf{h}_{E}\in\mathbb{R}^{m\times d}$ with $d$ being the hidden dimension.
\\
\textbf{Autoregressive decoder} in Figure \ref{figure:model} (a) conditions the current step on the encoded context and previously predictions to overcome the mismatch of sequence length. It calculates step by step
\[h_{D}^{i}=\mathrm{AR}(\mathrm{E}(a_{<i}), \mathbf{h}_{E})\in \mathbb{R}^{d}, i=0,\cdots,n+1,\]
\[\hat{a}_{i}=\mathrm{LogSoftmax}(h_{D}^{i})\in \mathbb{R}^{|\mathcal{V}_{\mathcal{A}}|}, i=0,\cdots,n+1,\]
and in the end, returns $\hat{\mathbf{a}}\in\mathbb{R}^{n\times|\mathcal{V}_{\mathcal{A}}|}$. The training is conducted as back-propgating $l(\mathbf{a}^{*}, \hat{\mathbf{a}})$. Note that $a_{0}^{*}=$ \texttt{BOS} and $a_{n+1}^{*}=$ \texttt{EOS} encourage the decoder to learn to begin and end the autoregression.
\\
\textbf{Non-autoregressive decoder} instead provides hidden states in one time. It is feasible to apply techniques of non-autoregressive machine translation. However, one of the primary issues solved by that is the uncertainty of the target sequence length. When it comes to state-action prediction, thanks to the flexibility at the sequence-level, we are allowed to design actions on purpose to eliminate such uncertainty. Specifically, we enforce action sequences to be of fixed length. On this basis, we propose D2 as shown in Figure \ref{figure:model} (b). To address the misalignment of sequence length between state and action, we insert a fully connected feed-forward network between the encoder and decoder\textsubscript{0}.
\[\mathrm{FFN}(\mathbf{h}_{E})=(\mathbf{h}_{E}^{\mathbf{T}}W + b)^{\mathbf{T}} \in\mathbb{R}^{n\times d}\]
where $W\in\mathbb{R}^{m\times n}$ and $b\in\mathbb{R}^{d \times n}$ transform the length dimension from $m$ to $n$ so as to project $\mathbf{h}_{E}$ into $\mathbf{h}_F\in\mathbb{R}^{n\times d}$. The alignment of the sequence length allows us to trivially pass $\mathbf{h}_{F}$ to decoder\textsubscript{0}.
\[\mathbf{h}_{D_{0}}=\mathrm{NAR}_{0}(\mathbf{h}_{F}, \mathbf{h}_{E}) \in \mathbb{R}^{n\times d}\]
\[\hat{\mathbf{a}}^{0}=\mathrm{LogSoftmax(\mathbf{h}_{D_{0}})} \in\mathbb{R}^{n\times|\mathcal{V}_{\mathcal{A}}|}\]
For a clear comparison with the autoregressive decoder, we make minimal changes to the structure and keep modeling the dependence between two contiguous steps through decoder\textsubscript{1}. To elaborate, we shift $\mathbf{a}^{0}$ one position to the right as $\acute{\mathbf{a}}^{0}$ by appending $a_{0}^{*}$ at the beginning and remove $a^{0}_{n}$ to maintain the sequence length. After that, we continue to feed $\acute{\mathbf{a}}^{0}$ to decoder\textsubscript{1}.
\[\mathbf{h}_{D_{1}}=\mathrm{NAR}_{1}(\mathrm{E}(\acute{\mathbf{a}}^{0}), \mathbf{h}_{E}) \in \mathbb{R}^{n\times d}\]
\[\hat{\mathbf{a}}^{1}=\mathrm{LogSoftmax(\mathbf{h}_{D_{1}})} \in\mathbb{R}^{n\times|\mathcal{V}_{\mathcal{A}}|}\]
At last, we conduct backpropagation with respect to the loss summation $l(\mathbf{a}^{*}, \hat{\mathbf{a}}^{0})\oplus l(\mathbf{a}^{*}, \hat{\mathbf{a}}^{1})$. Conventional seq2seq architectures are often equipped with intermediate modules such as a full attention distribution over the encoded context \citep{DBLP:journals/corr/BahdanauCB14}, which is omitted in the above formulation for simplicity. In the implementation, we always assume to train decoder\textsubscript{0} and decoder\textsubscript{1} separately to increase the model capacity, yet weight sharing is possible.

\subsection{Trajectory augmentation}

IL suffers from distribution shift and error accumulation \citep{ross2011reduction}. An agent's mistakes can easily put it into a state that the expert demonstrations do not involve and the agent has never seen during training. This also means errors can add up, so the agent drifts farther and farther away from the demonstrations. To tackle this issue, we propose TA that expands the expert demonstrations and actively exposes shifted states to the agent. We accomplish this by diverting intermediate states and consider them as initial states for TG. An example is offered in Figure \ref{figure:method} (b).

Given expert states $\mathbf{s}^{*}_{t\leq T}$ and corresponding actions $\mathbf{a}^{*}_{t\leq T}$, we utilize the divide-and-conquer technique to (1) break down the chain of state generation $\mathbf{s}^{*}_{t}\xmapsto[]{\mathbf{a}^{*}_{t}}\mathbf{s}^{*}_{t+1}$ into two by either executing $\mathbf{a}^{*}_{t}$ to stay on the current path or skipping $\mathbf{a}^{*}_{t}$ to branch the current path; (2) recursively calling this process until reaching the goal state $\mathbf{s}^{*}_{T}$; (3) merge intermediate states from branches and return from bottom to top in the end. As illustrated in Algorithm \ref{alg:traj_aug}, we collect a set of shifted states
\[\mathbf{S}^{\prime} = \mathrm{TA}(\emptyset, \mathbf{s}^{*}_{1}, \mathbf{s}^{*}_{t\leq T}, \mathbf{a}^{*}_{t\leq T}, \mathcal{E}),\]
regard them as initial states paired with the same goal state to produce extra trajectories
\[\tau^{\prime} = \underset{\mathbf{s}^{*} \in \mathbf{S}^{*}}{\cup}\mathrm{TG}(\mathbf{s}^{*}, \mathbf{s}^{*}_{T}, \mathcal{E}, \mathbf{E}),\]
and finally yield the augmented expert demonstrations $\mathcal{T}^{*}\cup\mathcal{T}^{\prime}$ after looping through $\mathcal{X}$.

TA is advantageous because it (i) only exploits existing expert demonstrations to preserve the i.i.d assumption; (ii) is universally applicable to our proposed paradigm without a dependency on the downstream task; (iii) does not need domain knowledge, labeling work, and further evaluation.

\section{Experiments} \label{sec:exp}

\begin{algorithm}[t]
    \small
    \caption{Trajectory Augmentation ($\mathrm{TA}$)}
    \label{alg:traj_aug}
    \renewcommand{\algorithmicrequire}{\textbf{Input:}}
    \renewcommand{\algorithmicensure}{\textbf{Output:}}
    \begin{algorithmic}[1]
        \Require
            States $\mathbf{S}$, state $\mathbf{s}_{t}$, expert states $\mathbf{S}^{*}$, actions $\mathbf{A}$, and environment $\mathcal{E}$.
        \Ensure
            Augmented states $\mathbf{S}$.
        \If{$|\mathbf{A}| > 1$}
            \State $\mathbf{a}_{t}\gets\mathbf{A}\mathrm{.pop}(0)$
            \State $\mathbf{s}_{t+1} \gets \mathcal{E}(\mathbf{s}_{t}, \mathbf{a}_{t})$
            \State $\mathbf{S} \gets \mathbf{S} \cup \mathrm{TA}(\mathbf{S}, \mathbf{s}_{t+1}, \mathbf{S}^{*}, \mathbf{A}, \mathcal{E})$ \Comment{Execute action}
            \State $\mathbf{A}\gets\mathrm{Update}(\mathbf{A}, \mathbf{s}_{t}, \mathbf{s}_{t+1})$
            \State $\mathbf{S} \gets \mathbf{S} \cup \mathrm{TA}(\mathbf{S}, \mathbf{s}_{t}, \mathbf{S}^{*}, \mathbf{A}, \mathcal{E})$ \Comment{Skip action}
        \ElsIf{$\mathbf{s}_{t} \notin \mathbf{S}^{*}$}
            \State $\mathbf{S} \gets \mathbf{S} \cup [\mathbf{s}_{t}]$ \Comment{Merge shifted state}
        \EndIf
        \State \Return $\mathbf{S}$
    \end{algorithmic}
\end{algorithm}

We adapt recurrent inference to our paradigm and evaluate them across AE benchmarks.

\subsection{Setup}

\begin{table*}[!htbp]
    \centering
    \resizebox{\textwidth}{!}{
        \begin{tabular}{ccccccccc}
        \toprule
        \multicolumn{3}{c}{\textbf{AOR} ($N=10$, $L=5$, $D=10$K)} & \multicolumn{3}{c}{\textbf{AES} ($N=100$, $L=5$, $D=10$K)} & \multicolumn{3}{c}{\textbf{AEC} ($N=10$, $L=5$, $D=10$K)} \\
        \cmidrule(r){1-3}
        \cmidrule(r){4-6}
        \cmidrule(r){7-9}
        \textbf{Train/Valid/Test} & \textbf{Train TA} & \textbf{Traj. Len.} & \textbf{Train/Valid/Test} & \textbf{Train TA} & \textbf{Traj. Len.} & \textbf{Train/Valid/Test} & \textbf{Train TA} & \textbf{Traj. Len.} \\
        \cmidrule(r){1-3}
        \cmidrule(r){4-6}
        \cmidrule(r){7-9}
        7,000/1,500/1,500 & 145,176 & 6 & 7,000/1,500/1,500 & 65,948 & 6 & 7,000/1,500/1,500 & 19,764 & 4 \\
        \bottomrule
        \end{tabular}
    }
    \caption{Data statistics of AE benchmarks.}
    \label{table:ap_data}
\end{table*}

\begin{table*}[t]
    \centering
    \resizebox{\textwidth}{!}{
        \begin{tabular}{llll}
        \toprule
        \textbf{Term} & \textbf{AOR} ($N=10$, $L=5$, $D=10$K) & \textbf{AES} ($N=100$, $L=5$, $D=10$K) & \textbf{AEC} ($N=10$, $L=5$, $D=10$K) \\
        \cmidrule(r){1-1}
        \cmidrule(r){2-2}
        \cmidrule(r){3-3}
        \cmidrule(r){4-4}
        Source $\mathbf{x}$ & 3 6 2 9 3 & 65 + ( 25 - 20 ) - ( 64 + 32 ) + ( 83 - 24 ) = ( - 25 + 58 ) & - 2 * + 4 10 + 8 / 8 = 8 \\ 
        Target $\mathbf{y}$ & - 3 - 6 / 2 + 9 = 3 & 65 + 5 - 96 + 59 = 33 & - 2 + 10 * 8 / 8 = 8 \\ 
        State $\mathbf{s}^{*}_{t}$ & - 3 - 6 / 2 9 3 & 65 + 5 - ( 64 + 32 ) + ( 83 - 24 ) = ( - 25 + 58 ) & - 2 + \textbf{4} 10 + 8 / 8 = 8 \\ 
        Action $\mathbf{a}^{*}_{t}$ & [POS\_6, +] & [POS\_4, POS\_8, 96] & [DELETE, POS\_3, POS\_3] \\
        Next State $\mathbf{s}^{*}_{t+1}$ & - 3 - 6 / 2 $\textbf{+}$ 9 3 & 65 + 5 - \textbf{96} + ( 83 - 24 ) = ( - 25 + 58 ) & - 2 + 10 + 8 / 8 = 8 \\ 
        Shifted State $\mathbf{s}^{\prime}_{t}$ & - 3 - 6 / 2 9 \textbf{=} 3 & 65 + 5 - ( 64 + 32 ) + \textbf{59} = ( - 25 + 58 ) & - 2 + 4 10 \textbf{*} 8 / 8 = 8 \\
        \bottomrule
        \end{tabular}
    }
    \vspace{-2mm}
    \caption{Examples from AE with specific $N$ for integer size, $L$ for the number of integers, and $D$ for data size.}
    \label{table:ae}
    \vspace{-2mm}
\end{table*}

\textbf{Data.} Arithmetic Operators Restoration (AOR) is a short-to-long editing to complete an array into a true equation. It is also a one-to-many task as an array can be completed as multiple true equations differently. Arithmetic Equation Simplification (AES) aims to calculate the parenthesized parts and keep the equation hold, resulting in a long-to-short and many-to-one editing. Arithmetic Equation Correction (AEC) targets to correct potential mistakes in an equation. Diverse errors perturb the equation, making AEC a mixed many-to-many editing. To align with the previous work, we follow the same data settings $N$, $L$, and $D$ for data generation, as well as the same action design for trajectory generation. The edit metric $\mathbf{E}$ for AOR and AEC is Levenshtein, while $\mathbf{E}$ for AES is a self-designed one (SELF) that instructs to replace tokens between two parentheses with the target token. Examples are presented in Table \ref{table:ae}. We refer readers to \citet{shi-etal-2020-recurrent} for an exhaustive explanation. As shown in Table \ref{table:ap_data}, the data splits are 7K/1.5K/1.5K for training, validation, and testing respectively.
\\
\textbf{Evaluation.} Sequence accuracy and equation accuracy are two primary metrics with token accuracy for a more fine-grained reference. In contrast to sequence accuracy for measuring whether an equation exactly matches the given label, equation accuracy emphasizes whether an equation holds, which is the actual goal of AE tasks. It is noted that there is no hard constraint to guarantee that all the predicted actions are valid. However, when the agent makes an inference mistake, the environment can refuse to execute invalid actions and keep the current state. This is also one of the beauties of reformulating text editing as a controllable MDP.
\\
\textbf{Baselines.} Recurrent inference (Recurrence) exhibits advantages over conventional end-to-end (End2end) and sequence tagging (Tagging) \citep{shi-etal-2020-recurrent} . However, for AES and AEC, it\footnote{\href{https://github.com/ShiningLab/Recurrent-Text-Editing}{\url{github.com/ShiningLab/Recurrent-Text-Editing}}} allows feeding training samples to a data generator and exposing more variants to models. These variants, as source samples paired with corresponding target samples, are used as the augmented dataset. This is impractical due to the strong dependency on domain knowledge. Given an input ``1 + \textbf{(2 + 2)} = 5'' and output ``1 + \textbf{4} = 5'' in AES, a variant ``1 + \textbf{(1 + 3}) = 5'' can be generated based on the knowledge \textbf{1 + 3 = 4}. Nevertheless, if this knowledge is not provided in the other training samples, the model should only know \textbf{2 + 2 = 4}.
\\
\textbf{Models.} As discussed, since the previously reported experiments are not practical, we re-run Recurrence source code for a more reasonable baseline (Recurrence*) that only has access to the fixed training set. Meanwhile, in our development environment, we reproduce Recurrence* within the proposed paradigm according to the compatibility in between. The encoder-decoder architecture inherits the same recurrent network as the backbone with long short-term memory units \citep{hochreiter1997long} and an attention mechanism \citep{luong-etal-2015-effective}. The dimension of the bidirectional encoder is 256 in each direction and 512 for both the embedding layer and decoder. We apply a dropout of 0.5 to the output of each layer \citep{srivastava2014dropout}. This provides us a standard autoregressive baseline AR, as well as a more powerful AR* after increasing the number of encoder layers from 1 to 4. On the one hand, to construct a non-autoregressive baseline NAR, we replace the decoder of AR* with a linear layer that directly maps the context to a probability distribution over the action vocabulary. In addition, we add two more encoder layers to maintain a similar amount of trainable parameters. On the other hand, replacing the decoder of AR* with D2 leads to our model NAR*. We strictly unify the encoder for a fair comparison regarding the decoder. Model configurations are shared across AE tasks for a comprehensive assessment avoiding particular tuning against any of them.
\\
\textbf{Training.} We train on a single NVIDIA Titan RTX with a batch size of 256. We use the Adam optimizer \citep{DBLP:journals/corr/KingmaB14} with a learning rate of $10^{-3}$ and an $\ell_2$ gradient clipping of $5.0$ \citep{10.5555/3042817.3043083}. A cosine annealing scheduler helps manage the training process and restarts the learning every 32 epochs to get it out of a potential local optimum. We adopt early stopping to wait for a lower validation loss until there are no updates for 512 epochs \citep{prechelt1998early}. Teacher forcing with a rate of $0.5$ spurs up the training process \cite{williams1989learning}. In AES and AEC, the adaptive loss weighting guides the model to adaptively focus on particular action tokens in accordance with the training results. Reported metrics attached with standard deviation are the results of five runs using random seeds from [0, 1, 2, 3, 4].

\subsection{Results} \label{sec:result}

\begin{table*}[t]
    \centering
    \resizebox{\textwidth}{!}{
        \begin{tabular}{lcccccccc}
        \toprule
        \multirow{2}{*}{\textbf{Method}} & \multicolumn{3}{c}{\textbf{AOR} ($N=10$, $L=5$, $D=10$K)} & \multicolumn{2}{c}{\textbf{AES} ($N=100$, $L=5$, $D=10$K)} & \multicolumn{3}{c}{\textbf{AEC} ($N=10$, $L=5$, $D=10$K)} \\
        \cmidrule(r){2-4}
        \cmidrule(r){5-6}
        \cmidrule(r){7-9}
         & \textbf{Tok. Acc. \%} & \textbf{Seq. Acc. \%} & \textbf{Eq. Acc. \%} & \textbf{Tok. Acc. \%} & \textbf{Eq. Acc. \%} & \textbf{Tok. Acc. \%} & \textbf{Seq. Acc. \%} & \textbf{Eq. Acc. \%} \\
        \cmidrule(r){1-1}
        \cmidrule(r){2-4}
        \cmidrule(r){5-6}
        \cmidrule(r){7-9}
        End2end & $-$ & $-$ & $29.33$ & $84.60$ & $25.20$ & $88.08$ & $57.27$ & $57.73$ \\
        Tagging & $-$ & $-$ & $51.40$ & $87.00$ & $36.67$ & $84.46$ & $46.93$ & $47.33$ \\
        Recurrence & $-$ & $-$ & $58.53$ & $98.63$ & $87.73$ & $83.64$ & $57.47$ & $58.27$ \\
        \cmidrule(r){1-1}
        \cmidrule(r){2-4}
        \cmidrule(r){5-6}
        \cmidrule(r){7-9}
        Recurrence* & $60.30\pm1.30$ & $27.31\pm1.33$ & $56.73\pm1.33$ & $79.82\pm0.37$ & $22.28\pm0.52$ & $82.32\pm0.56$ & $41.72\pm0.74$ & $42.13\pm0.75$ \\
        AR & $61.85\pm0.51$ & $28.83\pm1.14$ & $59.09\pm0.95$ & $88.12\pm2.37$ & $37.05\pm6.57$ & $\mathbf{82.61\pm0.53}$ & $45.81\pm0.36$ & $46.31\pm0.31$ \\
        AR* & $62.51\pm0.62$ & $\mathbf{30.85\pm0.41}$ & $61.35\pm0.33$ & $99.27\pm0.32$ & $93.57\pm2.91$ & $82.29\pm0.39$ & $45.99\pm0.49$ & $46.35\pm0.52$ \\
        NAR & $59.72\pm0.70$ & $24.16\pm1.16$ & $51.64\pm1.97$ & $83.87\pm1.60$ & $29.49\pm2.51$ & $80.28\pm0.76$ & $44.91\pm1.71$ & $45.40\pm1.78$ \\
        NAR* & $\mathbf{62.81\pm0.89}$ & $30.13\pm1.31$ & $\mathbf{61.45\pm1.61}$ & $\mathbf{99.51\pm0.13}$ & $\mathbf{95.67\pm0.93}$ & $81.82\pm0.68$ & $\mathbf{45.97\pm1.07}$ & $\mathbf{46.43\pm1.10}$ \\
        \cmidrule(r){1-1}
        \cmidrule(r){2-4}
        \cmidrule(r){5-6}
        \cmidrule(r){7-9}
        AR \texttt{\Large+}TA & $62.35\pm0.61$ & $32.28\pm0.67$ & $63.56\pm1.06$ & $88.05\pm1.20$ & $38.39\pm3.45$ & $\mathbf{83.94\pm0.42}*$ & $49.36\pm1.23$ & $49.83\pm1.21$ \\
        AR* \texttt{\Large+}TA & $62.58\pm0.63$ & $33.01\pm1.31$ & $65.73\pm1.38$ & $99.44\pm0.27$ & $95.24\pm2.38$ & $83.39\pm0.74$ & $48.95\pm0.65$ & $49.47\pm0.73$ \\
        NAR \texttt{\Large+}TA & $61.30\pm0.86$ & $32.04\pm1.99$ & $63.75\pm2.08$ & $90.38\pm2.21$ & $47.91\pm8.18$ & $81.36\pm0.40$ & $48.01\pm1.07$ & $48.47\pm1.15$ \\
        NAR* \texttt{\Large+}TA & $\mathbf{63.48\pm0.38}^*$ & $\mathbf{34.23\pm0.92}^*$ & $\mathbf{67.13\pm0.99}^*$ & $\mathbf{99.58\pm0.15}^*$ & $\mathbf{96.44\pm1.29}^*$ & $82.70\pm0.42$ & $\mathbf{49.64\pm0.59}^*$ & $\mathbf{50.15\pm0.55}^*$ \\
        \bottomrule
        \end{tabular}
    }
    \vspace{-2mm}
    \caption{Evaluation results on AOR, AES, and AEC with specific $N$, $L$, and $D$. The token and sequence accuracy for AOR were not reported, thus we leave these positions blank here. With or without TA, our proposed NAR* achieves the best performance in terms of equation accuracy across the board.}
    \label{table:exp}
\end{table*}

\begin{figure}[t]
    \centering
    \includegraphics[width=\linewidth]{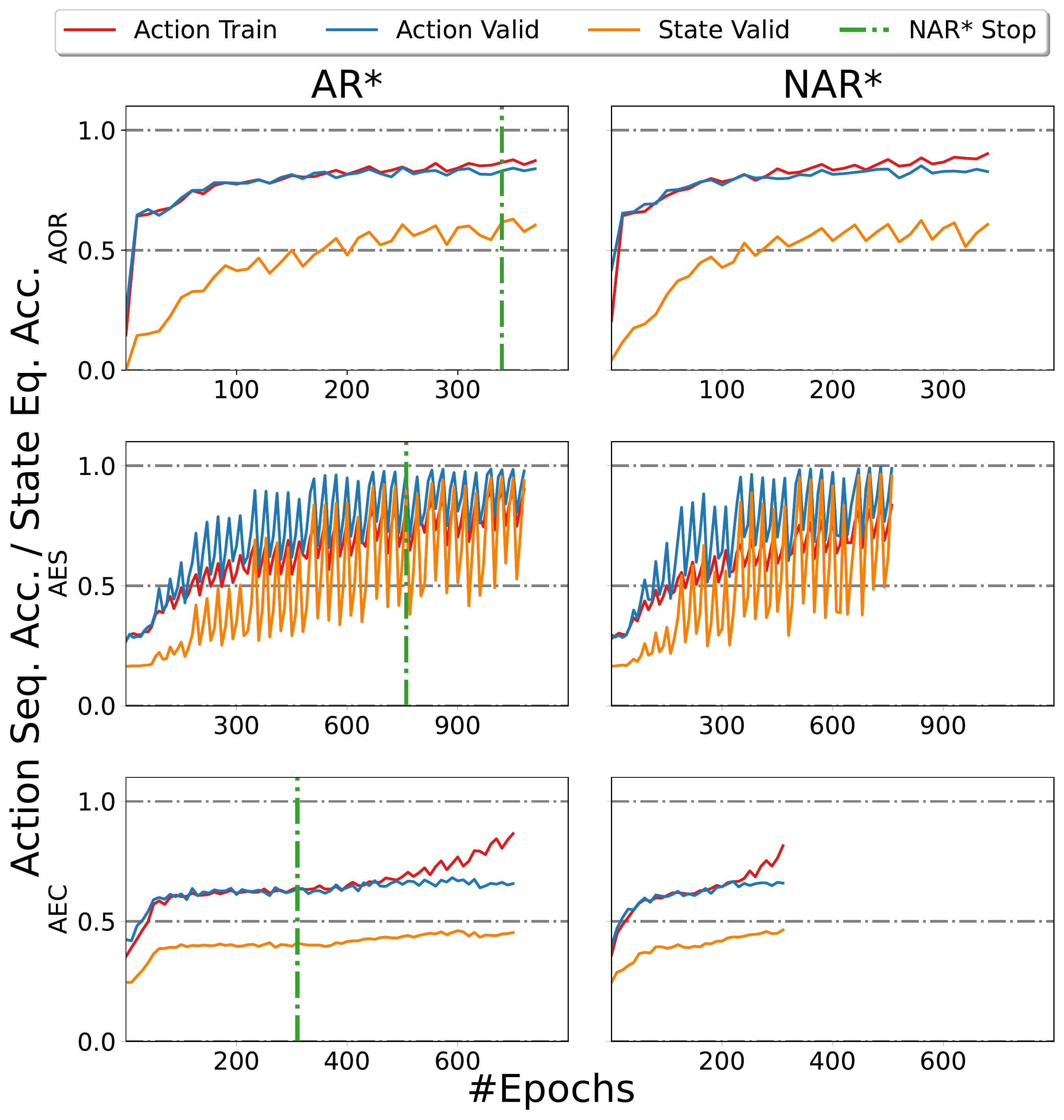}
    \vspace{-6mm}
    \caption{The learning curve of AR* (left column) and NAR* (right column) across AE tasks (rows). The red and blue lines represent the training on actions w.r.t sequence accuracy. The orange line stands for the validation on returned states w.r.t equation accuracy. The dashed line in green marks the earlier stop epoch of NAR* than that of AR* during training.}
    \label{figure:exp_lc}
    \vspace{-2mm}
\end{figure}

\textbf{Baselines.} As summarized in Table \ref{table:exp}, prohibiting the access of Recurrence to domain knowledge outcomes a fair baseline and significantly weakens Recurrence* in AES and AEC. We also would like to point out that, even in the same impractical setting, our NAR* can achieve around 99.33\% and 67.49\% for AES and AEC with respect to equation accuracy, which is still much higher than that (87.73\% and 58.27\% for AES and AEC) reported in the previous work. In AOR, a one-to-many editing, no augmented source sequence is retrieved from the target side. We confirm that the slight accuracy drop of Recurrence* in AOR results from bias through multiple tests. Although AR is our reproduction of Recurrence*, the overall advancement of AR over Recurrence* proves the goodness of our framework and implementation. Participation of added three encoder layers in AR* improves model capacity and thus contributes to higher accuracy. A simple linear header already enables NAR to parallel the decoding; nevertheless, it dramatically reduces performance, especially in AES.
\\
\textbf{Non-autoregressive.} What stands out is the dominance of NAR*, achieving $61.45\%$, $95.67\%$, and $46.43\%$ in terms of equation accuracy for AOR, AES, and AEC, separately. Particularly in AES, its better performance over AR* by more than $2.1\%$ equation accuracy underlines the success of NAR* in capturing the interdependencies among target tokens. Its superiority with respect to equation accuracy boosting by around $66.18\% $ over NAR highlights the contributions of D2 again.
\\
\textbf{Trajectory augmentation.} As expected, the incorporation of TA consistently promotes the accuracy of all models in our learning regime throughout AE tasks. Taking NAR as an example, training with TA brings it a substantial equation accuracy gain, remarkably up to $18.42\%$ in AES. Even more, it pushes the gap between NAR* and the other baselines. The most notable advance comes from AOR, where NAR* outperforms AR* by a substantial margin of $5.68\%$ equation accuracy. It appears that TA is more effective for non-autoregressive models than autoregressive ones.

\begin{figure}[t]
    \centering
    \includegraphics[width=\linewidth]{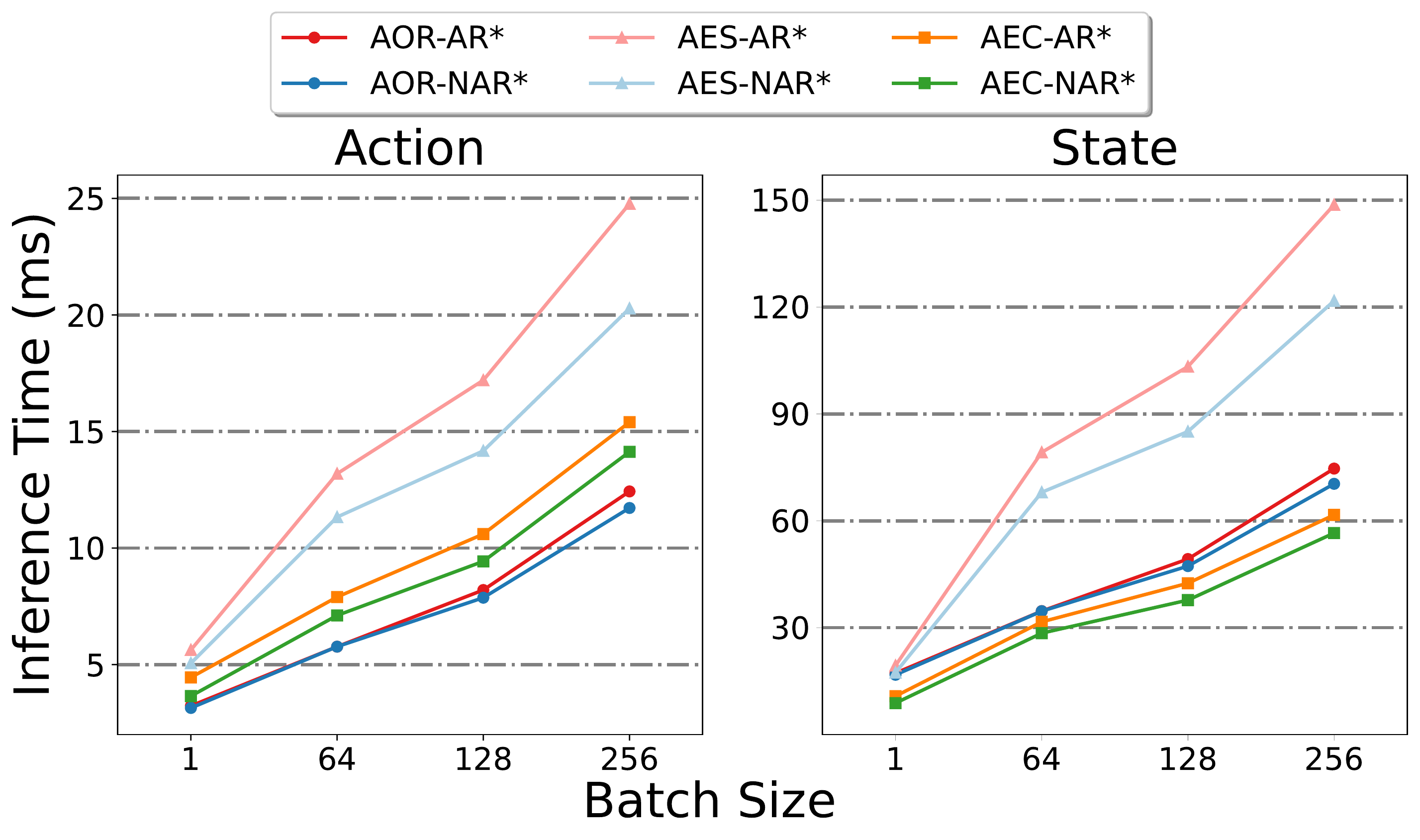}
    \vspace{-6mm}
    \caption{Inference time of AR* and NAR* to predict action (left) and return state (right) across AE tasks.}
    \label{figure:exp_infer}
    \vspace{-2mm}
\end{figure}

\section{Analysis}

We conduct extensive sensitivity analyses to better illustrate and understand our methods.

\subsection{Efficiency}

From the learning curve (Figure \ref{figure:exp_lc}) and inference time (Figure \ref{figure:exp_infer}) of AR* and NAR* in AE, in addition to a higher accuracy, we find NAR* needs less number of training epochs to converge and trigger the early stopping. The periodic fluctuation of the learning curve is the consequence of using a scheduler. When it comes to inference, NAR* saves much time for every step of action determination and ends up returning the edited state faster. As AR* and NAR* share exactly the same encoder structure, we conclude that D2 contributes to the advanced efficiency.

\begin{table}[t]
    \begin{center}
        \resizebox{\linewidth}{!}{
            \begin{tabular}{cclcc}
                \toprule
                \textbf{Design} & \textbf{Action Sequence} & \textbf{Method} & \textbf{Tok. Acc. \%} & \textbf{Eq. Acc. \%} \\
                \cmidrule(r){1-2}
                \cmidrule(r){3-3}
                \cmidrule(r){4-5}
                \multirow{4}{*}{\#1} & \multirow{4}{*}{[\texttt{Pos.\textsubscript{L}}, \texttt{Pos.\textsubscript{R}}, \texttt{Tok.}]} & AR* & $99.27\pm0.32$ & $93.57\pm2.91$\\
                & & NAR* & $\mathbf{99.51\pm0.13}$ & $\mathbf{95.67\pm0.93}$ \\
                \cmidrule(r){3-3}
                \cmidrule(r){4-5}
                & & AR* \texttt{\Large+}TA & $99.44\pm0.27$ & $95.24\pm2.38$ \\
                & & NAR* \texttt{\Large+}TA & $\mathbf{99.58\pm0.15}^*$ & $\mathbf{96.44\pm1.29}^*$ \\
                \cmidrule(r){1-2}
                \cmidrule(r){3-3}
                \cmidrule(r){4-5}
                \multirow{4}{*}{\#2} & \multirow{4}{*}{[\texttt{Pos.\textsubscript{L}}, \texttt{Tok.}, \texttt{Pos.\textsubscript{R}}]} & AR* & $99.08\pm0.93$ & $92.35\pm7.21$ \\
                & & NAR* & $\mathbf{99.50\pm0.27}$ & $\mathbf{95.55\pm2.28}$ \\
                \cmidrule(r){3-3}
                \cmidrule(r){4-5}
                & & AR* \texttt{\Large+}TA & $99.52\pm0.29$ & $95.68\pm2.49$ \\
                & & NAR* \texttt{\Large+}TA & $\mathbf{99.54\pm0.20}^*$ & $\mathbf{95.97\pm1.64}^*$ \\
                \cmidrule(r){1-2}
                \cmidrule(r){3-3}
                \cmidrule(r){4-5}
                \multirow{4}{*}{\#3} & \multirow{4}{*}{[\texttt{Tok.}, \texttt{Pos.\textsubscript{L}}, \texttt{Pos.\textsubscript{R}}]} & AR* & $98.06\pm0.79$ & $83.79\pm6.25$ \\
                & & NAR* & $\mathbf{99.53\pm0.14}$ & $\mathbf{95.99\pm0.81}$ \\
                \cmidrule(r){3-3}
                \cmidrule(r){4-5}
                & & AR* \texttt{\Large+}TA & $98.43\pm0.49$ & $87.29\pm3.70$ \\
                & & NAR* \texttt{\Large+}TA & $\mathbf{99.61\pm0.06}^*$ & $\mathbf{96.55\pm0.46}^*$ \\
                \bottomrule
            \end{tabular}
        }
    \end{center}
    \vspace{-2mm}
    \caption{Evaluation of AR* and NAR* in AES across three action designs that vary from each other by token order. They directs to the same operation with \texttt{Pos.\textsubscript{L}}/\texttt{Pos.\textsubscript{R}}/\texttt{Tok.} denoting left parenthesis/right parenthesis/target token.}
    \label{table:analysis_ad}
\end{table}

\subsection{Action design}

Due to the liberty of sequence generation, the same operation can be represented as different action sequences. In AES, the operation, instructing to substitute tokens between left and right parentheses with the required token, can fit the three action designs in Table \ref{table:analysis_ad}, where \texttt{Pos.\textsubscript{L}}, \texttt{Pos.\textsubscript{R}}, and \texttt{Tok.} denote the positions of two parentheses and the target token. Design \#1 is the default one. A simple swap of action tokens offers designs \#2 and \#3.

AR* severely suffers such perturbation, causing an equation accuracy decline by $9.78\%$ in \#3. Contrastly, NAR* holds around its results and even slightly improves to $95.99\%$ in \#3. Despite the joining of TA, AR* still goes down from $95.24\%$ in \#1 to $87.29\%$ in \#3, while NAR* stays nearly consistent across three designs. It is reasonable that AR* is sensitive to the order of action tokens because the position information helps the inference of the target token. This also reflects that NAR* can catch the position information but with little dependence on token order. Such robustness allows greater freedom of action design.

\begin{table}[t]
    \begin{center}
        \resizebox{\linewidth}{!}{
            \begin{tabular}{lllcc}
                \toprule
                \textbf{Edit Metric} $\mathbf{E}$ & $\bm{T_{\max}}$ & \textbf{Method} & \textbf{Tok. Acc. \%} & \textbf{Eq. Acc. \%} \\
                \cmidrule(r){1-1}
                \cmidrule(r){2-2}
                \cmidrule(r){3-3}
                \cmidrule(r){4-5}
                \multirow{2}{*}{SELF} & \multirow{2}{*}{6} & AR* & $99.27\pm0.32$ & $93.57\pm2.91$\\
                 & & NAR* & $\mathbf{99.51\pm0.13}$ & $\mathbf{95.67\pm0.93}$ \\
                \cmidrule(r){1-1}
                \cmidrule(r){2-2}
                \cmidrule(r){3-3}
                \cmidrule(r){4-5}
                \multirow{2}{*}{Levenshtein} & \multirow{2}{*}{31} & AR* & $\mathbf{69.53\pm2.29}$ & $\mathbf{18.37\pm0.70}$ \\
                 & & NAR* & $67.58\pm0.87$ & $17.93\pm0.07$ \\
                \bottomrule
            \end{tabular}
        }
    \end{center}
    \vspace{-2mm}
    \caption{Evaluation of AR* and NAR* trained with edit metrics SELF and Levenshtein in AES. $\bm{T_{\max}}$ refers to the maximum length of expert trajectories.}
    \label{table:analysis_to_aes}
\end{table}

\begin{figure}[t]
    \centering
    \includegraphics[width=\linewidth]{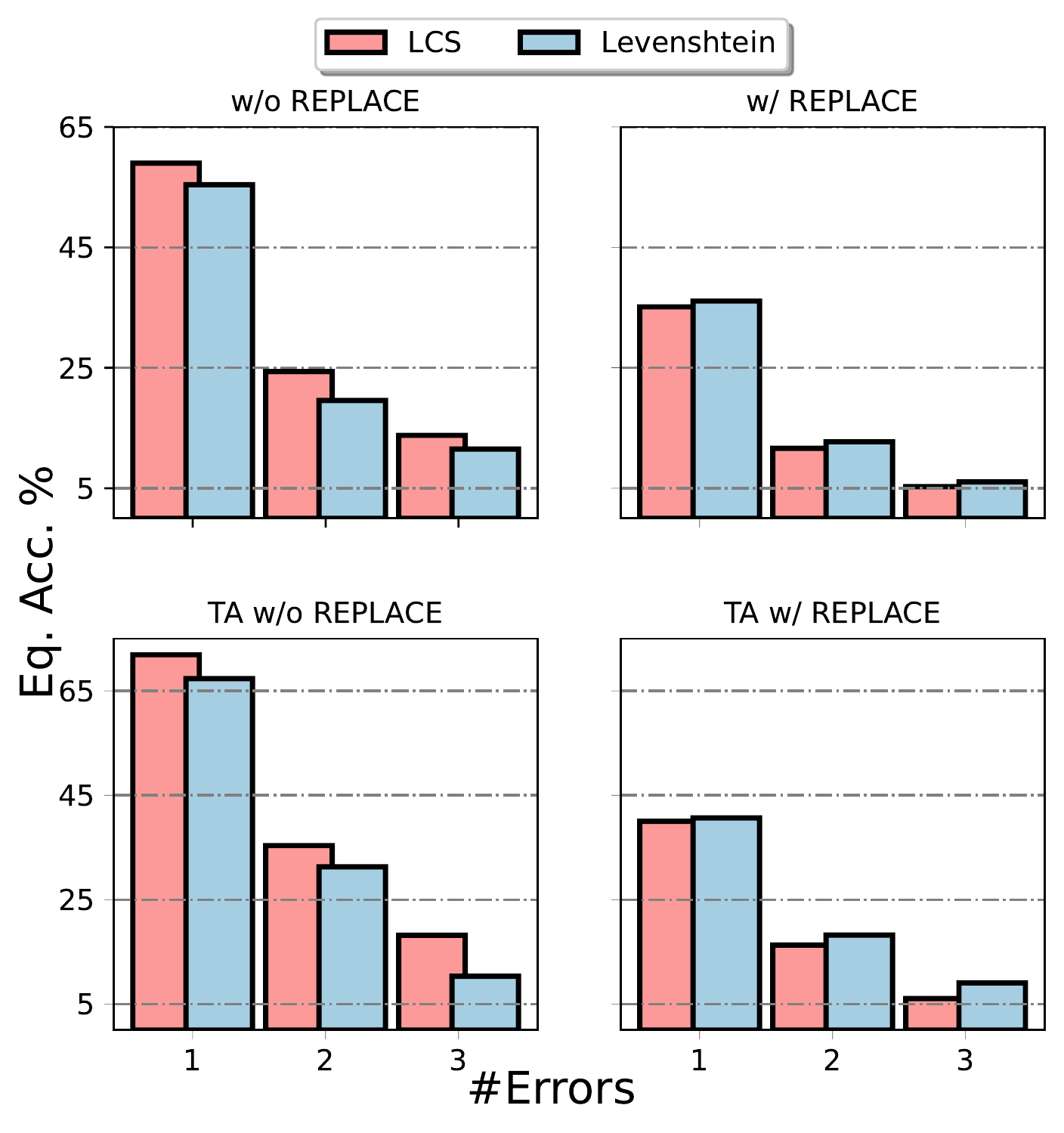}
    \vspace{-6mm}
    \caption{Evaluation of NAR* trained with edit metrics LCS and Levenshtein in AEC. Results are grouped by two trajectory lengths caused by whether the policy involves \texttt{REPLACE}.}
    \label{figure:analysis_to_aec}
    \vspace{-2mm}
\end{figure}

\begin{table*}[t]
    \centering
    \resizebox{\textwidth}{!}{
        \begin{tabular}{lcccccccc}
        \toprule
        \multirow{2}{*}{\textbf{Decoder}} & \multicolumn{3}{c}{\textbf{AOR} ($N=10$, $L=5$, $D=10$K)} & \multicolumn{2}{c}{\textbf{AES} ($N=100$, $L=5$, $D=10$K)} & \multicolumn{3}{c}{\textbf{AEC} ($N=10$, $L=5$, $D=10$K)} \\
        \cmidrule(r){2-4}
        \cmidrule(r){5-6}
        \cmidrule(r){7-9}
         & \textbf{Tok. Acc. \%} & \textbf{Seq. Acc. \%} & \textbf{Eq. Acc. \%} & \textbf{Tok. Acc. \%} & \textbf{Eq. Acc. \%} & \textbf{Tok. Acc. \%} & \textbf{Seq. Acc. \%} & \textbf{Eq. Acc. \%} \\
        \cmidrule(r){1-1}
        \cmidrule(r){2-4}
        \cmidrule(r){5-6}
        \cmidrule(r){7-9}
        Linear & $61.84\pm0.94$ & $28.55\pm1.57$ & $57.72\pm1.55$ & $99.41\pm0.26$ & $95.01\pm2.01$ & $81.35\pm0.92$ & $42.47\pm1.85$ & $42.81\pm1.87$ \\
        Decoder\textsubscript{0} & $61.78\pm0.83$ & $28.20\pm1.57$ & $58.36\pm1.58$ & $99.24\pm0.23$ & $93.49\pm2.03$ & $80.84\pm0.66$ & $43.97\pm1.82$ & $44.32\pm1.82$ \\
        Shared D2 & $61.74\pm0.71$ & $28.68\pm0.94$ & $58.05\pm1.01$ & $99.28\pm0.24$ & $93.85\pm2.14$ & $81.38\pm1.04$ & $43.64\pm2.03$ & $44.09\pm2.02$ \\
        D2 (NAR*) & $\mathbf{62.81\pm0.89}$ & $\mathbf{30.13\pm1.31}$ & $\mathbf{61.45\pm1.61}$ & $\mathbf{99.51\pm0.13}$ & $\mathbf{95.67\pm0.93}$ & $\mathbf{81.82\pm0.68}$ & $\mathbf{45.97\pm1.07}$ & $\mathbf{46.43\pm1.10}$ \\
        \cmidrule(r){1-1}
        \cmidrule(r){2-4}
        \cmidrule(r){5-6}
        \cmidrule(r){7-9}
        Linear \texttt{\Large+}TA & $61.41\pm0.28$ & $31.75\pm0.93$ & $63.15\pm0.96$ & $99.42\pm0.17$ & $95.08\pm1.47$ & $81.54\pm0.66$ & $46.79\pm2.26$ & $47.33\pm2.30$ \\
        Decoder\textsubscript{0} \texttt{\Large+}TA & $62.50\pm1.24$ & $32.48\pm1.87$ & $64.47\pm1.88$ & $99.47\pm0.13$ & $95.33\pm1.13$ & $82.02\pm0.40$ & $46.80\pm2.04$ & $47.32\pm1.91$ \\
        Shared D2 \texttt{\Large+}TA & $61.64\pm0.87$ & $31.21\pm0.34$ & $62.77\pm0.85$ & $99.53\pm0.12$ & $95.91\pm1.25$ & $81.80\pm0.47$ & $47.23\pm1.07$ & $47.61\pm1.14$ \\
        D2 (NAR*) \texttt{\Large+}TA & $\mathbf{63.48\pm0.38}^*$ & $\mathbf{34.23\pm0.92}^*$ & $\mathbf{67.13\pm0.99}^*$ & $\mathbf{99.58\pm0.15}^*$ & $\mathbf{96.44\pm1.29}^*$ & $\mathbf{82.70\pm0.42}^*$ & $\mathbf{49.64\pm0.59}^*$ & $\mathbf{50.15\pm0.55}^*$ \\
        \bottomrule
        \end{tabular}
    }
    \vspace{-2mm}
    \caption{Evaluation of agents equipped with same encoders but different decoders on AE benchmarks.}
    \label{table:analysis_model}
\end{table*}

\subsection{Trajectory optimization}

A better edit metric $\mathbf{E}$ often means a smaller action vocabulary space $|\mathcal{V}_{\mathcal{A}}|$, shorter trajectory length $T_{\max}$, and, therefore, an easier IL. Taking AES as an instance, a SELF-action, replacing tokens enclosed in parentheses with the target one, actually is the compression of several Levenshtein-actions including multiple deletions and one substitution. Although either can serve as an expert policy, SELF causes a much shorter $T_{\max}$ as indicated in Table \ref{table:analysis_to_aes}. The change from SELF to Levenshtein brings on a longer $T_{\max}$ and consequently a significant performance gap of $75.2\%$ and $77.74\%$ for AR* and NAR* in terms of equation accuracy. Doing one edit in 31 steps rather than 6 undoubtedly raises the difficulty of the imitation game.

As one more exploration, we introduce Longest Common Subsequence (LCS) as an alternative $\mathbf{E}$ to AEC. Token replacement is not allowed in LCS but in Levenshtein. A replacement action has to be decomposed as one deletion and one insertion in LCS. From this, LCS has a small $|\mathcal{V}_{\mathcal{A}}|$, while Levenshtein has a shorter $T_{\max}$. We train NAR* with these two and report in Figure \ref{figure:analysis_to_aec}. For a clear comparison, the test set is divided into two groups. In \texttt{w/o REPLACE}, both yield the same $T_{\max}$, but, in \texttt{ w/ REPLACE}, Levenshtein takes a shorter $T_{\max}$. In the former, LCS exceeds Levenshtein with or without TA. In the latter, the opposite is true, where Levenshtein outperforms LCS under the same condition. This support our assumption at the beginning that an appropriate $\mathbf{E}$, leading to a small $|\mathcal{V}_{\mathcal{A}}|$ and a short $T_{\max}$, is conducive to IL, suggesting trajectory optimization an interesting future work.

\subsection{Dual decoders}

As an ablation study, we freeze the encoder of NAR* and vary its decoder to reveal the contributions of each component in D2. As listed in Table \ref{table:analysis_model}, replacing the decoder with a linear layer leads to Linear and removing the second decoder from NAR* results in Decoder\textsubscript{0}. Moreover, sharing the parameters between two decoders of NAR* gives the Shared D2. All of them can parallel the decoding process. We then borrow the setup of Section \ref{sec:exp} and test them on AE.

Among four decoders, NAR* dominates three imitation games. The performance decrease caused by shared parameters is more significant than expected. Besides the reason that saved parameters limit the model capacity, another potential one is the input mismatch of two decoders. The input of decoder\textsubscript{0} is the projected context from the linear layer after the encoder, yet that of decoder\textsubscript{1} is the embedded prediction from the embedding layer. When incorporating TA, we find the same trend persists. The gap between NAR* and the others is even more apparent. Since they share the same encoder, such a gap clarifies the benefits of D2.

\section{Conclusion}

We reformulate text editing as an imitation game defined by an MDP to allow action design at the sequence-level. We propose D2, a non-autoregressive decoder for state-action learning, coupled with TG for data translation and TA for distribution shift alleviation. Achievements on AE benchmarks evidence the advantages of our methods in performance, efficiency, and robustness. Sequence-level actions are arguably more controllable, interpretable, and similar to human behavior. Turning tasks into games that agents feel more comfortable with sheds light on future studies in the direction of reinforcement learning in the application of text editing. The involvement of a reward function, the optimization of the trajectories, the design of sequence-level actions, and their applications in more practical tasks, to name a few, are interesting for future work. Suggesting text editing as a new testbed, we hope our findings will shed light on future studies in reinforcement learning applying to natural language processing.

\section*{Limitations}
Each time the state is updated, the agent can get immediate feedback on the previous action and thus a dynamic context representation during the editing. This also means that the encoder (e.g., a heavy pretrained language model) will be called multiple times to refresh the context matrix. Consequently, as the trajectory grows, the whole task becomes slow even though we have paralleled the decoding process. Meanwhile, applying our methods in more realistic editing tasks (e.g., grammatical error correction) remains a concern and needs to be explored in the near future.


\section*{Acknowledgements}
We gratefully appreciate Che Wang (Watcher), Yichen Gong, and Hui Xue for sharing their pearls of wisdom. We also would like to express our special thanks of graitude to Yingying Huo for the support, as well as EMNLP anonymous reviewers for their constructive feedback. This work was supported by Shining Lab and Alibaba Group.

\bibliography{anthology,custom}
\bibliographystyle{acl_natbib}



\end{document}